\documentclass[letterpaper, 10 pt, conference]{ieeeconf}  
\usepackage{xcolor}
\usepackage{hyperref}
\usepackage{amsfonts}
\usepackage{amsmath}
\usepackage{cite}
\usepackage{adjustbox}
\usepackage{hyperref}

\usepackage{etoolbox,siunitx,booktabs,threeparttable,multirow,array,graphicx}
\usepackage[normalem]{ulem}

\usepackage{cite}
\usepackage{graphicx} 
\usepackage{amsmath,amssymb,amsfonts}
\usepackage{graphicx}
\usepackage{textcomp}
\usepackage{xcolor}
\usepackage{hyperref}
\usepackage{booktabs}
\usepackage{subcaption}
\usepackage{multirow}
\usepackage{svg}
\usepackage{float}
\usepackage{algorithm}
\usepackage{algpseudocode}

\newcommand{\model}{T-CFM}
\newcommand{\smodel}{T-CFM }

\newcommand{\mrs}{CADENCE}

\newcommand{\oldmodel}{GrAMMI }

\robustify\bfseries
\robustify\uline
\robustify\tnote

\newcommand{\feedback}[1]{\textcolor{black}{#1}}
\newcommand{\fed}[1]{\textcolor{black}{#1}}
\newcommand{\rev}[1]{\textcolor{black}{#1}}

\IEEEoverridecommandlockouts                              

\usepackage{amssymb}  

\title{\LARGE \bf
Efficient Trajectory Forecasting and Generation with Conditional Flow Matching
}

 \author{Sean Ye$^{1}$ and Matthew C. Gombolay$^{1}$
 \thanks{*This work is supported in part by Naval Research Laboratory (NRL) under grant number N00173-21-1-G009 and the National Science Foundation under grant CNS-2219755,}
 \thanks{$^{1}$All authors are associated with the Institute of Robotics and Intelligent Machines (IRIM), Georgia Institute of Technology, Atlanta, GA, USA.}%
 \thanks{Corresponding Author: Sean Ye,
 		{\tt\small seancye@gatech.edu}}
 }

\begin{document}

\maketitle
\thispagestyle{empty}
\pagestyle{empty}



\begin{abstract}
Trajectory prediction and generation are crucial for autonomous robots in dynamic environments. While prior research has typically focused on either prediction or generation, our approach unifies these tasks to provide a versatile framework and achieve state-of-the-art performance. While diffusion models excel in trajectory generation, their iterative sampling process is computationally intensive, hindering robotic systems' dynamic capabilities. We introduce Trajectory Conditional Flow Matching (T-CFM), a novel approach using flow matching techniques to learn a solver time-varying vector field for efficient, fast trajectory generation. T-CFM demonstrates effectiveness in adversarial tracking, real-world aircraft trajectory forecasting, and long-horizon planning, outperforming state-of-the-art baselines with 35\% higher predictive accuracy and 142\% improved planning performance. Crucially, T-CFM achieves up to 100$\times$ speed-up compared to diffusion models without sacrificing accuracy, enabling real-time decision making in robotics. Codebase: https://github.com/CORE-Robotics-Lab/TCFM

\end{abstract}


\section{INTRODUCTION}

\feedback{Robots of the future will require fast and accurate trajectory forecasting techniques to navigate complex, dynamic environments and interact with other agents safely and efficiently. Trajectory forecasting deals with the problem of estimating an agent's future behavior while trajectory generation deals with planning feasible paths for an agent to follow. These techniques are crucial for various robotics applications, such as autonomous driving \cite{salzmann2020trajectron}, multi-robot coordination \cite{wu2023adversarial}, and social navigation \cite{che2020efficient}. By generating long-horizon plans and accurately predicting the future trajectories of dynamic agents, robots can make better decisions and adapt to changing conditions in real-time.}

\begin{figure}[ht]
\centering
    \includegraphics[width=\columnwidth]{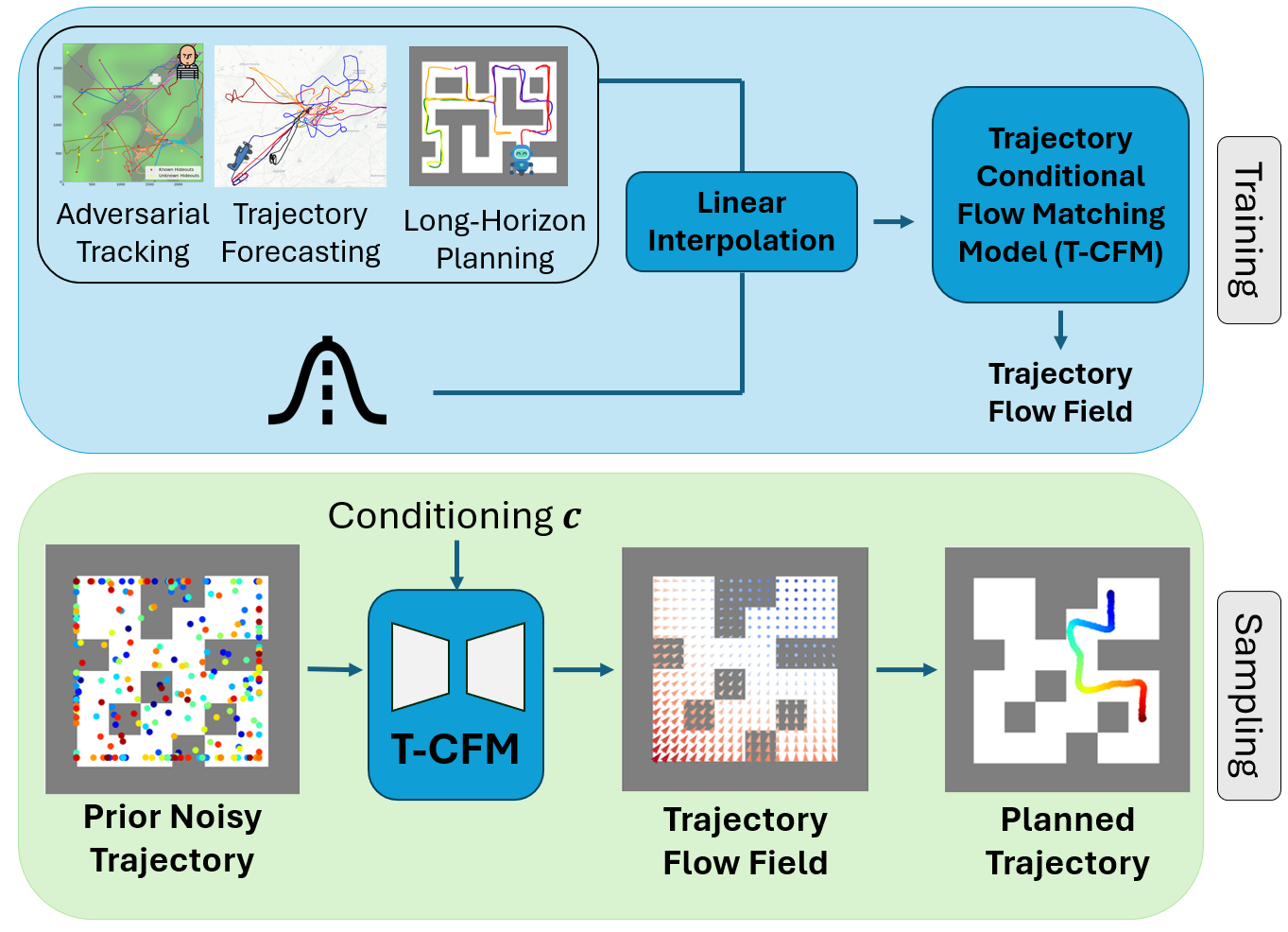}
    \caption{Trajectory Conditional Flow Matching (T-CFM) is our novel trajectory prediction and generation framework. The model is capable of generating trajectories in a single step, outperforming prior generative modeling work by learning a time-varying vector field to sample trajectories.}
    \label{fig:small_overview}
    \vspace{-5mm}
\end{figure}

In recent years, deep learning approaches have achieved impressive results on trajectory forecasting benchmarks by learning complex patterns and distributions from large datasets. \feedback{Compared to traditional methods such as Kalman Filters \cite{leven2009unscented} and Particle Filters \cite{djuric2008target}, learning-based methods excel at tasks where models of an agent's behavior are unknown or hard to predict.} In particular, generative models such as variational autoencoders (VAEs) and generative adversarial networks (GANs) have shown promise in modeling the inherent multimodality and stochasticity in agent behaviors. More recently, denoising diffusion probabilistic models (DDPMs) \cite{song2021scorebased} have emerged as a powerful class of generative models, demonstrating superior performance in sample quality and diversity across various domains. However, a key limitation of diffusion models is their slow sampling speed, which hinders their real-time applicability in robotics. \feedback{VAEs and GANs are fast but struggle with multimodal sample quality compared to diffusion models.}

In this paper, we introduce a novel trajectory forecasting and generation framework that employs flow matching \cite{lipman2023flow}, a method that transforms between data distributions using a learned time-varying vector field. Our technique, named Trajectory Conditional Flow Matching (T-CFM), \feedback{maintains the sample quality of diffusion models while generating samples an order of magnitude faster by circumventing the iterative sampling approach used in diffusion models.} 

\feedback{We demonstrate the efficacy of T-CFM on three robotics tasks, \fed{shown in} Figure \ref{fig:dataset_overview}. In the adversarial tracking scenario, autonomous pursuing agents predict the future trajectories of an adversarial evader. We also showcase T-CFM's performance on a real-world aircraft trajectory prediction dataset, which has important implications for the development of autonomous aerial robots. Accurate trajectory prediction enables these robots to avoid collisions and coordinate with other aircrafts. Finally, we apply T-CFM to long-horizon planning in complex 2D maze environments \cite{fu2020d4rl}, demonstrating our model's ability to generate long-horizon plans, which is crucial for robot navigation.}

We show that T-CFM outperforms state-of-the-art baselines, including diffusion models, in terms of predictive accuracy of generated trajectories and sample quality of generated plans. Notably, our approach can generate high-quality trajectory samples with as few as one sampling step, leading to significant speed-ups compared to diffusion models, without sacrificing performance. 


\noindent \textbf{Contributions:} Our key contributions are three-fold. 
\begin{itemize}
    \item We propose T-CFM, a novel flow matching framework for conditional trajectory forecasting and generation that is both accurate and efficient. To the best of our knowledge, we are the first to apply flow matching to trajectory prediction and trajectory planning tasks.
    \item We demonstrate state-of-the-art performance on three challenging robotics tasks: adversarial tracking, aircraft trajectory forecasting, and long-horizon planning, achieving up to in 35\% increase in prediction accuracy and 142\% in planning performance. 
    \item \feedback{T-CFM achieves significant sampling time speed-ups compared to prior generative modeling approaches, reducing sampling time by up to 100×. Our framework is versatile and can generate high-quality trajectories using as few as one sampling step or multi-step sampling when needed.}
\end{itemize}




    
    

    


\section{Related Works}
\label{sec:related_work}

Traditionally, trajectory forecasting and trajectory planning have occupied very different and distinct avenues of robotics research. However, with the advent of deep learning, these tasks have become increasingly intertwined. In this section, we review traditional methods for both trajectory forecasting and generation (Section \ref{sec:symbolic}). Then we describe why learning-based approaches can \fed{address the key limitations} of traditional methods and introduce flow matching and generative modeling. 

\subsection{Symbolic Methods for Trajectory Forecasting \& Planning}
\label{sec:symbolic}

\paragraph{Target Tracking}
Target tracking is a well-studied problem in the robotics community \cite{bar2004estimation}, with numerous applications, including surveillance \cite{grocholsky2006}, crowd monitoring \cite{tokekar2014}, and wildlife monitoring \cite{dunbabin2012}. Traditional methods for target tracking and trajectory prediction, such as Kalman Filters \cite{chen2000mixture, leven2009unscented} and Particle Filters \cite{djuric2008target, rao2013visual}, have been widely used in various in applications like autonomous navigation, object tracking in video surveillance, robotics, and radar systems. However, their performance degrades when faced with sparse observations, lack of accurate target behavior models, and long prediction horizons \cite{li2003}. In our work, we address these limitations by leveraging a flow matching-based approach that learns to model the target's behavior from data, allowing for accurate predictions even in sparse observation settings and over long horizons.

\paragraph{Planning and Navigation}
Traditional methods for path planning, such as RRT* and PRM* \cite{karaman2011sampling}, are widely used in environments with a known representation of the world. However, they can be computationally expensive for large state spaces and require a priori knowledge, which may not always be available or can change dynamically. Learning-based approaches, such as ours, offer the promise of generalization and the ability to provide solutions in dynamic environments. Our approach learns to generate feasible trajectories directly from data, eliminating the need for explicit maps and enabling fast planning in complex, dynamic environments.

\subsection{Learning-Based Approaches}

\paragraph{Supervised Learning Methods for Trajectory Prediction}
Recent works in trajectory prediction for various domains, such as aircraft navigation (FlightBERT) \cite{guo23flightbert}, social navigation  \cite{che2020efficient}, and autonomous driving \cite{salzmann2020trajectron}, have employed supervised learning methods. These approaches often utilize autoregressive models and log-likelihood based training to learn predictive models from data. Graph-based Adversarial Modeling with Mutual Information (GrAMMI) \cite{ye2023learning} is a recent framework that explicitly models a multimodal distribution using a combination of a Gaussian Mixture Model regularized by mutual information. While effective in certain scenarios, these models can struggle with capturing long-horizon multi-modal distributions, which are common in real-world trajectory data. We leverage a generative modeling approach rather than a \rev{discriminative} one, enabling more diverse multimodal outputs for accurate prediction in complex real-world scenarios.

\paragraph{Generative Modeling}
 Generative modeling techniques, which have shown great success in computer vision tasks \cite{song2021scorebased}, provide a promising avenue to augment trajectory prediction by learning to model complex, multimodal distributions. Recently, diffusion-based probabilistic models \cite{song2021scorebased} have dominated many generative modeling tasks. Diffusion models generate samples by iteratively denoising a Gaussian distribution, allowing them to capture complex, multimodal distributions. Diffuser \cite{janner2022planning} and \rev{Motion Planning Diffusion \cite{carvalho2023mpd}}, are diffusion based approaches for learning trajectories, representing the state of the art in long horizon planning and offline reinforcement learning. Constrained Agent-based Diffusion for Enhanced Multi-Target Tracking (CADENCE) \cite{ye2023diffusion}, similarly extends the Diffuser framework for Adversarial Tracking and modeling multi-agent behaviors. The main drawback of diffusion models is their iterative denoising process, which can be computationally slow, limiting their real-time applicability.

As an efficient alternative to diffusion models, flow matching techniques \cite{lipman2023flow, tong2023improving} learn a generative model using ordinary differential equations (ODEs) instead of stochastic differential equations (SDEs). This formulation allows for faster sampling while maintaining the ability to model complex distributions. To the best of our knowledge, our work is the first to apply flow matching techniques for learning trajectories for prediction and planning tasks.

\color{black}
\section{\feedback{Preliminaries}}
Trajectories play a crucial role in various robotics domains as they represent the behavior and evolution of an agent over time. Formally, we define a trajectory $\tau$ as a sequence of states ${s^1, s^2, \ldots, s^T}$, where $s^t \in \mathcal{S}$ represents the state of the agent at time horizon step $t$, and $\mathcal{S}$ is the state space. In some cases, trajectories may also include actions, represented as $\tau = \{(s^1, a^1), (s^2, a^2), \ldots, (s^T, a^T)\}$, where $a^t \in \mathcal{A}$ is the action taken by the agent at time step $t$, and $\mathcal{A}$ is the action space.

\begin{figure*}[htbp]
\vspace*{1.5mm}
\centering
    \includegraphics[width=0.82\textwidth]{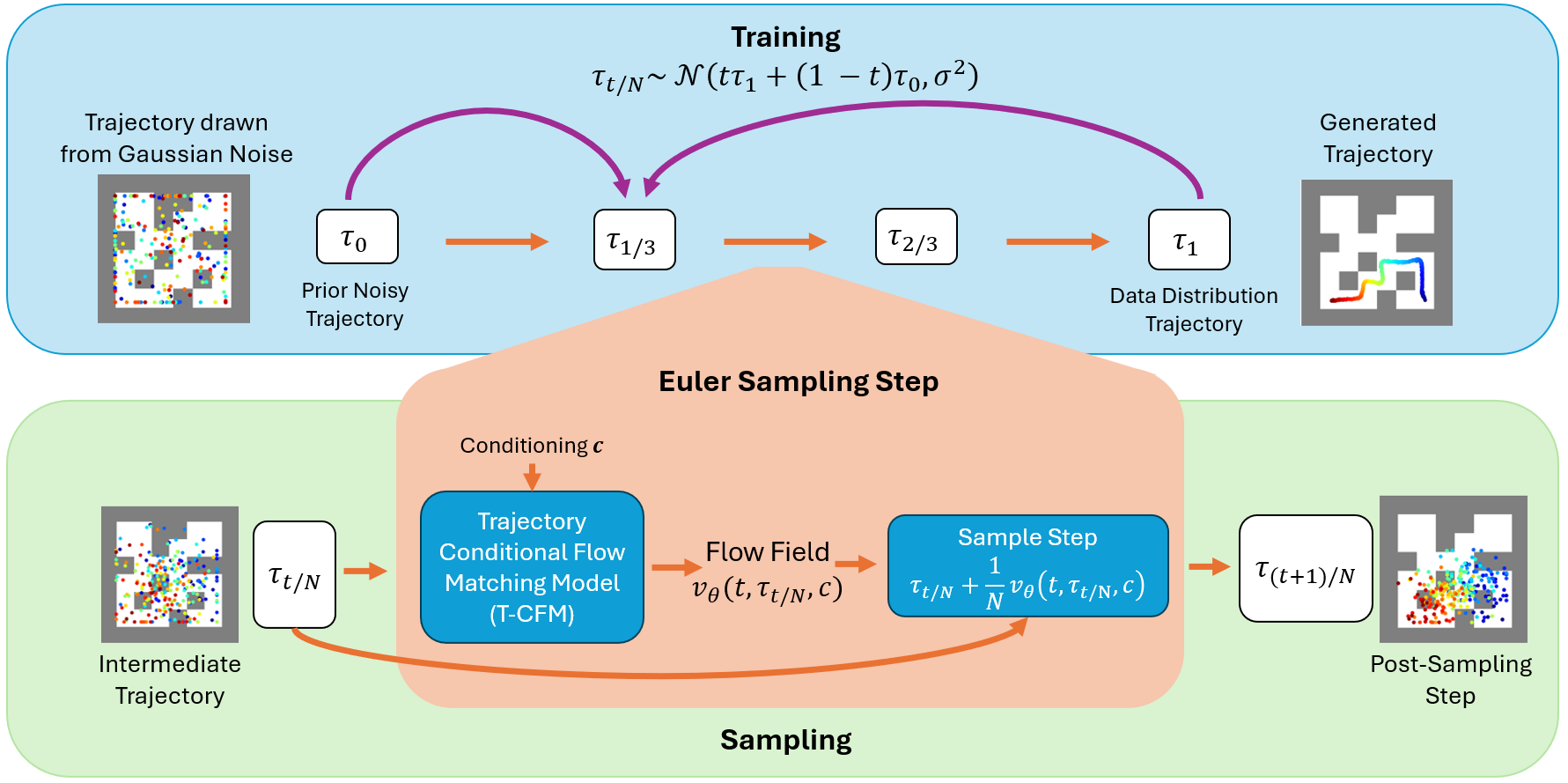}
    \caption{Overview of Trajectory Conditional Flow Matching. The flow matching formulation defines intermediate trajectories as a linear combination between the prior noise distribution ($\tau_0$) and data distribution ($\tau_1)$. The sampling procedure then utilizes the learned flow field generated by the model to create samples.}
    \label{fig:overview}
\end{figure*}

By modeling trajectories in this general form, we can develop methods for trajectory forecasting and planning that are applicable across different domains. In trajectory forecasting, the goal is to predict an agent's future trajectory given its past states and additional context information, $c$. This can be formalized as learning a conditional distribution $p_\theta(\tau^{t+1:T} | \tau^{1:t}, c)$, where $\tau^{1:t}$ represents the observed trajectory up to time $t$, $\tau^{t+1:T}$ represents the future trajectory to be predicted, and $c$ represents any additional context. Similarly, in adversarial tracking, we can use the same framework to predict an adversary's $(\tau)$ future trajectory given prior detection or environmental information, $c$.

In long-horizon planning, the goal is to generate a trajectory that leads an agent from an initial state to a goal state. This can be formalized as drawing a trajectory sample from $p_\theta(\tau|c)$, where the context information, $c$ are the start and goal states. Rather than a trajectory optimization problem, this formulation allows us to view the problem as a conditional generation task, where we sample trajectories from a learned distribution conditioned on the desired start and goal states.

\section{Method}
\label{sec:method}

    

In this section, we describe our flow matching formulation and how we model our problem. The goal primary goal is to generate trajectories $\tau$ given the conditioning factor $c$. Trajectories are defined simply as a sequence of states or a sequence of states and actions. 

\subsection{Flow Matching Formulation}
\feedback{The goal of flow matching, similar to diffusion models, is to learn a process that can generate samples through an iterative process that lies in the data distribution. To do this, we model the starting random Gaussian noise distribution as $q(\tau_0)$ and the trajectory data distribution as $q(\tau_1)$. We refer to these distributions as $q_0, q_1$ where the generative modeling task is to transform $q_0$ to $q_1$.}

\feedback{To learn a model that can transform $q_0$ to $q_1$}, we model a time-varying vector field $u: [0,1] \times \mathbb{R}^d \rightarrow \mathbb{R}^d$ and a probability path $p: [0, 1] \times \mathbb{R}^d \rightarrow \mathbb{R}^{d+}$. The vector field $u$ is defined by an ordinary differential equation (Equation \ref{eq:ode}).

\begin{equation}
d\tau = u_t(\tau) dt
\label{eq:ode}
\end{equation}

\feedback{Intuitively, the vector field defines the direction and magnitude to push each sample such that a sample from $q_0$ arrives at its corresponding location in $q_1$ by following the probability path $p$ over time.}

We aim to approximate the true vector field $u$ using a neural network represented by $v_\theta(t, \tau)$, where $v_\theta(t, \tau)$ defines a time-dependent vector field parameterized by weights, $\theta$. The flow matching objective is to minimize the difference between the predicted vector field $v_\theta(t, \tau)$ and the true vector field $u_t(\tau)$, as expressed in Equation \ref{eq:fm}.

\begin{equation}
\min_{\theta} \mathbb{E}_{t,\tau \sim p_t(\tau)} \left\| v_\theta(t, \tau) - u_t(\tau) \right\|^2
\label{eq:fm}
\end{equation}

However, this objective is intractable as there is no closed form representation for the true vector field $u_t(\tau)$. Instead, prior work \cite{lipman2023flow, tong2023improving, albergo2023building} has proposed to estimate the conditional form of the vector field $u_t(\tau|z)$ which is conditioned on a random variable $z$. In our work, we use the formulation where $q(z) = q(\tau_0) q(\tau_1)$, \feedback{meaning $z$ captures the starting and ending points of the trajectory.}

\feedback{We assume a Gaussian flow between $\tau_0$ and $\tau_1$ with standard deviation $\sigma$ and model the probability path $p_t(\tau|z)$ and vector field $u_t(\tau|z)$ as shown in Equations \ref{eq:pu}. Equation \ref{eq:pt} defines the probability path as a Gaussian distribution centered at a linear interpolation between $\tau_0$ and $\tau_1$ at time $t$. In the top portion of Figure \ref{fig:overview}, we show a visualization of the linear interpolation used to generate intermediate trajectories between $\tau_0$ and $\tau_1$. Equation \ref{eq:ut} defines the target vector field simply as the difference vector pointing from the starting point $\tau_0$ to the end point $\tau_1$.}

\vspace{-1.1mm}
\begin{subequations}
\label{eq:pu}
\begin{equation}
p_t(\tau|z) = \mathcal{N} \left( \tau | t \tau_1 + (1 - t) \tau_0, \sigma^2 \right)
\label{eq:pt}
\end{equation}
\vspace{-6.5mm}
\begin{equation}
u_t(\tau|z) = \tau_1 - \tau_0.
\label{eq:ut}
\end{equation}
\end{subequations}
\vspace{-2mm}

With this formulation, we now have a computable target vector field that we can regress our neural network to. Algorithm \ref{alg:cfm} summarizes the training steps:
\begin{enumerate}
    \item Draw a starting trajectory $\tau_0$ from the Gaussian noise distribution $q(\tau_0)$ and a random timestep $t$ from a uniform distribution (Lines \ref{line:draw} - \ref{line:timestep}).
    \item Draw a ground truth end trajectory $\tau_1$ and conditioning factor $c$ from the dataset (Line \ref{line:draw_two}).
    \item Compute the intermediate trajectory $\tau$ at time $t$ by linearly interpolating between $\tau_0$ and $\tau_1$ (Line \ref{line:compute}).
    \item Match the vector field $v_\theta(t, \tau)$ predicted by the neural network to the target vector field $u_t(\tau|z) = \tau_1 - \tau_0$ (Lines \ref{line:target} - \ref{line:update_t}).
\end{enumerate}
\feedback{By repeating these steps and updating the neural network weights to minimize the difference between the predicted and target vector fields, the model learns to approximate the true time-dependent vector field that transforms samples from the starting noise distribution to the data distribution.}

\begin{algorithm}[htbp]
\caption{Conditional Flow Matching Training}
\begin{algorithmic}[1]
\Require Dataset $\mathcal{D}$, computable $u_t(x|z)$ and network $v_{\theta}(t, \tau, c)$.
\While{Training}
    \State $\tau_0 \sim q(\tau_0)$; \Comment Draw source trajectory
    \label{line:draw}
    \State $t \sim U(0, 1)$ \Comment Draw timestep
    \label{line:timestep}
    \State $\tau_1, c \sim \mathcal{D}$ \Comment Draw target trajectory and conditioning
    \label{line:draw_two}
    \State $\tau \sim p_t(\tau|z) = \mathcal{N} \left( \tau | t \tau_1 + (1 - t) \tau_0, \sigma^2 \right)$ \Comment Eq. \ref{eq:pt}
    \label{line:compute}
    \State $u_t(\tau|z) = \tau_1 - \tau_0$ \Comment Eq. \ref{eq:ut} 
    \label{line:target}
    \State $\mathcal{L}_{CFM}(\theta)=||v_{\theta}(t, \tau, c) - u_t(\tau|z)||^2$ \Comment Loss
    \label{line:update}
    \State $\theta = \theta + \alpha \nabla_\theta \mathcal{L}_{CFM}(\theta)$ \Comment Update Model
    \label{line:update_t}
\EndWhile
\end{algorithmic}
\label{alg:cfm}
\end{algorithm}

The neural network model used to parameterize $v_\theta(t, \tau, c)$ is a 1D Convolutional Temporal U-Net based on prior diffusion work \cite{janner2022planning, ye2023diffusion}. 1D convolutions slide over the time dimension of the input trajectory $\tau$, capturing temporal patterns and dependencies without being autoregressive. This allows for efficient parallel processing of the entire trajectory. The model also incorporates Feature-Wise Linear Modulation (FiLM) Layers \cite{perez2018film} to condition the model with relevant context information, $c$. \rev{For each domain, we will describe the context vector ($c$) used (Section \ref{sec:eval})}. By using the same base architecture as prior diffusion work, we demonstrate that our training methodology generates better models irrespective of model parameter count and architecture.

\subsection{Sampling}

Given a trained flow model $v_\theta(t, \tau, c)$, the sampling procedure utilizes an ODE solver to recover the solution to Equation \ref{eq:ode}. We can denote the solution of the ODE with $\phi_t(\tau)$, where $\phi_0(\tau) = \tau$ and $\phi_t(\tau)$ is the transformation of our trajectory $\tau$ transported along the vector field from time $0$ to time $t$. In Algorithm \ref{alg:sample}, we show the sampling procedure using the Euler method (Line \ref{line:euler}) but any off the shelf ODE solver can be used. In our experiments, we choose to use the Euler sampling method as the number of sampling steps is easily adjustable. \feedback{The bottom portion of Figure \ref{fig:overview} shows how the sampling procedure moves from a prior noisy trajectory $\tau_0$ to a trajectory that lies within the data distribution $\tau_1$. }

\fed{One key difference between the trajectory generation and planning tasks is the planning task requires constraints on the sampled trajectory.} We provide a formulation to constrain the generated trajectory $\tau$ to start at the current robot state and end at the desired goal state. For each sampling step, we set these states of the trajectory rather than interpolate from noise (Alg \ref{alg:sample}, Line \ref{line:inpaint}-\ref{line:inpaint_1}), where the horizon of the trajectory is denoted $h$. In this formulation, we allow the model to infill the trajectory to generate a cohesive plan.



\begin{algorithm}[htbp]
\caption{Euler Sampling}
\begin{algorithmic}[1]
\Require Samplable $q(\tau_0) = \mathcal{N}(0, I)$, trained flow network $v_{\theta}(t, \tau, c)$, number of sampling timesteps $N$
\Statex \textbf{Optional}: Start State $s^0$, End State $s^h$ 
\State $\tau_0 \sim q(\tau_0)$ \Comment Sample $\tau_0$
\For{$t = 1, ..., N$}
    \If{Planning}
        \State $\tau_{t/N}^0 \leftarrow s^0$ \Comment Set Start State
        \label{line:inpaint}
        \State $\tau_{t/N}^h \leftarrow s^h$ \Comment Set Goal State
        \label{line:inpaint_1}
    \EndIf
    \State $\tau_{(t+1)/N} \leftarrow \tau_{t/N} + \frac{1}{N} v_{\theta}(t, \tau_{t/N}, c)$
    \label{line:euler}
\EndFor
\end{algorithmic}
\label{alg:sample}
\end{algorithm}

\feedback{The key feature of parameterizing the probability flow and vector field through Equation \ref{eq:pu} is that it enables our model to learn \textit{straight} flows as compared to diffusion models. In diffusion models, the sampling process involves gradually denoising a Gaussian noise sample over many steps, following a complex path in the data space. This typically requires a large number of sampling steps to generate high-quality samples. In contrast, our flow matching approach learns a direct, straight path from the starting noise distribution to the target data distribution by modeling the probability path as a linear interpolation between the starting and ending points (Equation \ref{eq:pu}), encouraging the model to find the most efficient trajectory that matches the true data distribution.}

\feedback{Intuitively, the straight flows learned by our model can be thought of as a shortcut from the noise distribution to the data distribution. Instead of taking a meandering path through the data space, the model learns to follow a direct route guided by the target vector field, enabling our method to reduce the number of intermediate steps needed to generate high-quality trajectories. This straight path allows for faster sampling with fewer steps.}

\section{Evaluation and Domains}
\label{sec:eval}

\begin{figure}[t]
\vspace*{1.5mm}
\centering
    \includegraphics[width=0.83\columnwidth]{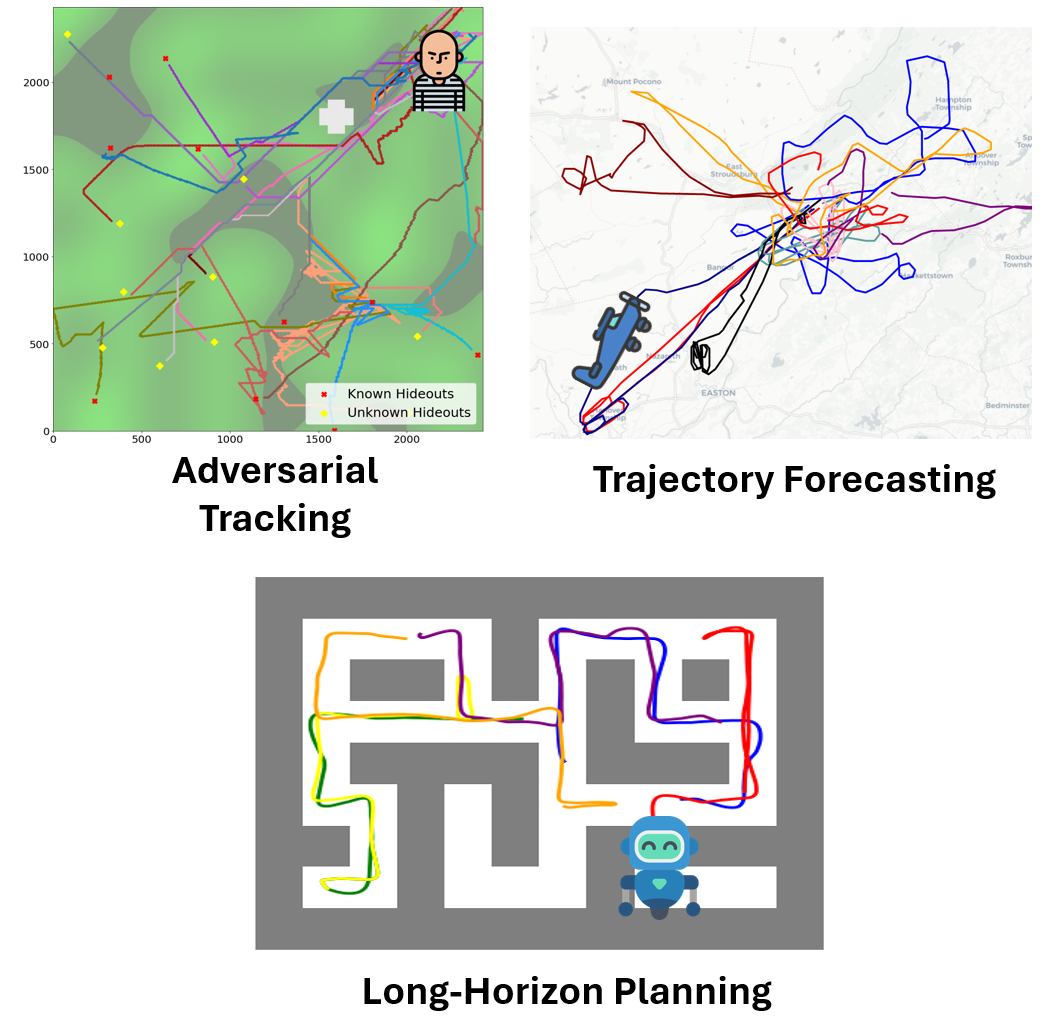}
    \caption{Trajectory Forecasting and Planning Domains: Our T-CFM framework is applicable many trajectory modeling tasks, with Adversarial Tracking, Trajectory Forecasting, and Long-Horizon Planning domains shown here. }
    \label{fig:dataset_overview}
    \vspace{-5mm}
\end{figure}

We test our model in three different tasks and domains: 1) Adversarial Tracking 2) Trajectory Forecasting, and 3) Long-Horizon Planning. These domains test our model's capability of generating accurate multimodal trajectory predictions and plans for robots to use. Visualizations for the training data and domains are shown in Figure \ref{fig:dataset_overview} and a \rev{summary of the states and context vectors used are in Table \ref{tab:domain_comparison}}.

\setlength{\tabcolsep}{3pt}
\begin{table}[ht] \centering
\begin{tabular}{l|ll}
Domain & State in Trajectory ($\tau$) & Context Vector ($c$)\\
\toprule
Adversarial Tracking & Position (x, y) & Historical detection \\
 & coordinates & information \\
\midrule
Aircraft Trajectory & Longitude, Latitude, & 5-minute history \\
 & Altitude & of past states \\
\midrule
Long-Horizon Planning & Position (x, y) & Start and goal \\
 & coordinates & states \\
\bottomrule
\end{tabular}
\caption{Summary of State and Context Vectors for Different Domains}
\vspace{-2mm}
\label{tab:domain_comparison}
\end{table}


\subsection{Adversarial Tracking}
Adversarial tracking aims to predict an adversary's future trajectory $\tau$ given past historical information, $c$. These domains are challenging due to the adversary's potential multiple strategies and the observers' often incomplete or sparse data. We assess our flow-matching tracking models using the Prison Escape scenarios, as introduced in previous work \cite{ye2023learning}.

\begin{figure*}[t]
\vspace*{1mm}
\centering
\begin{subfigure}{0.44\textwidth}
\centering
\includegraphics[width=\textwidth]{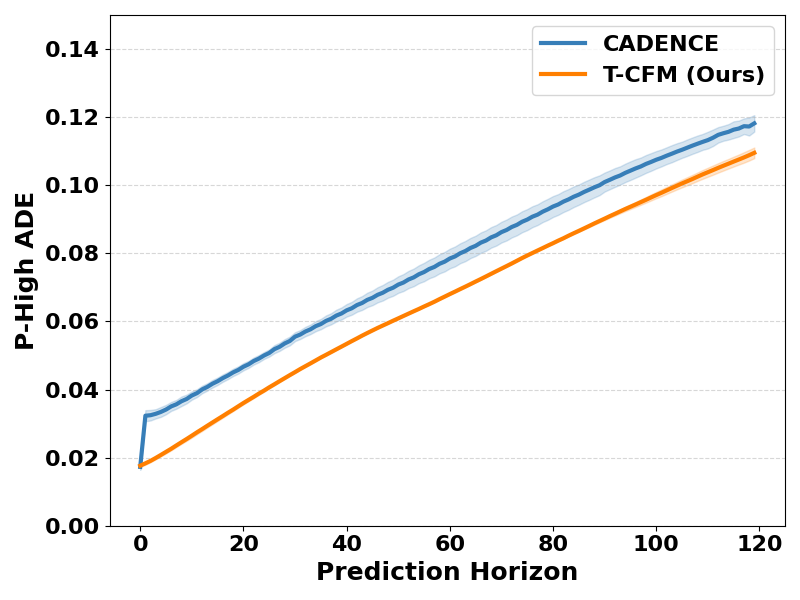}
\caption{ADE vs Prediction Time-Horizon}
\label{fig:cfm_result}
\end{subfigure}
\hfill
\begin{subfigure}{0.48\textwidth}
\centering
\includegraphics[width=\textwidth]{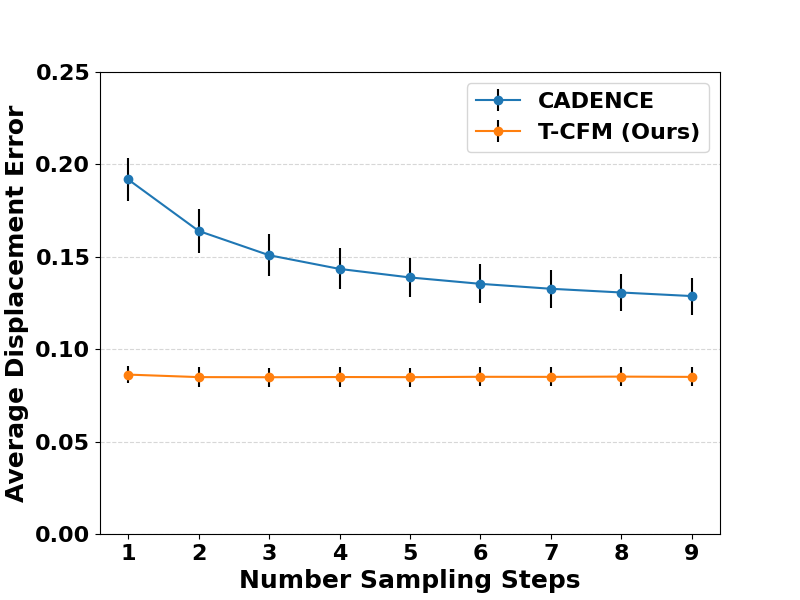}
\caption{ADE vs Total Sampling Steps}
\label{fig:adv_sample_speed}
\end{subfigure}
\caption{Comparison our model with the diffusion-based CADENCE model. Our method achieves better ADE on the entire prediction horizon (left) while also maintaining performance when the number of sampling steps is lowered (right).}
\label{fig:combined}
\end{figure*}

\feedback{The Prisoner Escape and Narco Traffic Interdiction simulations share similar pursuit-evasion dynamics, with tracking agents collaborating to locate and apprehend an adversary attempting to reach predetermined hideouts. The agents face the challenge of operating in large environments with sparse detections of the opponent. Key differences between the domains include the type of fog-of-war, agent capturing dynamics, and destination types. We refer the reader to prior work for more details \cite{ye2023learning}.}

For both scenarios, we utilize open-sourced datasets from prior work \cite{ye2023learning}. The Prison Escape scenario consists of three datasets (Prisoner-Low, Prisoner-Medium, Prisoner-High) with opponent detection rates of $12.9\%$, $44.0\%$, and $63.1\%$, respectively. The Narco Interdiction scenario uses two datasets with opponent detection rates of $13.8\%$ and $31.5\%$, adjusted by modifying the pursuit agents' detection radius. We evaluate our models using Average Displacement Error (ADE), which computes the average $l_2$ distance between each sampled trajectory and the ground truth trajectory over all timesteps.

\subsection{Aircraft Trajectory Forecasting}
To demonstrate the capabilities of our model on real data, we retrieved two years of data for a single Cessna aircraft from the OpenSky database \cite{schafer2014bringing}. Individual trajectories were extracted from the dataset, resulting in a total of 474 trajectories and a train/val/test split of 80/10/10\% was used. The Cessna was chosen to constrain the range of trajectories while maintaining significant variability, allowing for testing the multimodal performance of our algorithm. \feedback{The goal is to predict the future trajectory $\tau$ given the 5-minute history of past states $c$, which we use as our context vector.} We evaluate the forecasting performance using two common metrics: mean absolute error (MAE) and root mean square error (RMSE) for longitude, latitude, and altitude.


\subsection{Long-Horizon Imitation Learning - Maze2D}
\feedback{Learning to plan for long horizons is crucial for robots to navigate autonomously in complex domains.} The performance of our models in long-horizon planning is evaluated using the Maze2D environments \cite{fu2020d4rl}. In this task, the agent must traverse from a starting location to a goal location. The algorithm is tested on three maps of increasing difficulty: U-Maze, Medium, and Large. Following prior work, the performance is reported in terms of \textit{score}, which represents the agent's success in reaching the final goal. The score is normalized between 0 and 100 based on an expert policy. 

Two different evaluations are performed: single-task and multi-task. In the single-task evaluation, the goal location remains constant while in the multi-task setting, the goal location is randomly selected at the beginning of each episode. Training data consists of successful trajectories between randomly selected start and end goals.

\section{Results and Discussions}
\label{sec:results}

This section presents results and analysis for three tasks: Adversarial Tracking, Trajectory Forecasting, and Long-Horizon Planning. Three models were trained for each task using different random seeds.

\subsection{Adversarial Tracking}
Our approach is benchmarked against several state-estimation baselines including 1) VRNN \cite{chung2015recurrent}, 2) GRaMMI \cite{ye2023learning} and 3) CADENCE \cite{ye2023diffusion}. 

\subsubsection{Tracking Capabilities of Flow Matching}
\setlength{\tabcolsep}{3pt}
\newcommand{\ra}[1]{\renewcommand{\arraystretch}{5}}

\begin{table}[ht] \centering
\begin{tabular}{c|l|rrrrr}
\multicolumn{2}{l}{} & \multicolumn{5}{c}{\textbf{Prediction Horizon}} \\
\multicolumn{1}{l}{}             &                  & 0 min           & 30 min         & 60 min         & 90 min         & 120 min        \\
\toprule
\multirow{3}{*}{\rotatebox[origin=c]{90}{P-Low}} 
& Particle Filter &  0.120 & 0.148 & 0.161 & 0.171 & 0.183\\
& VRNN             & {0.106}  & 
                                 {0.093} & 
                                 {0.119} & 
                                 {0.146} & 
                                 {0.177} \\
                                 & \oldmodel & 0.060            & 0.080           & 0.110           & 0.154          & 0.163          \\
                                  & \mrs             & 0.057  & 0.077 & \textbf{0.100}   & \textbf{0.127} & 0.154 \\
                                  & \smodel (Ours)            & \textbf{0.055} & \textbf{0.076} & 0.101 & 0.128 & \textbf{0.153} \\

\midrule
\multirow{3}{*}{\rotatebox[origin=c]{90}{P-Med}} 
& Particle Filter & 0.099 & 0.129 & 0.141 & 0.152 & 0.163 \\
& VRNN             & {0.172}  & 
                                 {0.086} & 
                                 {0.110} & 
                                 {0.144} & 
                                 {0.167} \\ 
                                 & \oldmodel & 0.049           & 0.077          & 0.110           & 0.146          & 0.167          \\
                                 & \mrs             & 0.046 & 0.076 & 0.103 & 0.129 & 0.153 \\
                                 & \smodel (Ours)  & \textbf{0.030} & \textbf{0.058} & \textbf{0.088} & \textbf{0.118}& \textbf{0.146} \\

\midrule
\multirow{3}{*}{\rotatebox[origin=c]{90}{P-High}}  
& Particle Filter & 0.041 & 0.084 & 0.102 & 0.119 & 0.133 \\
& VRNN             & {0.105}  & 
                                 {0.059} & 
                                 {0.100} & 
                                 {0.117} & 
                                 {0.145} \\
                                 & \oldmodel & \textbf{0.015}        & 0.056          & 0.092          & 0.122          & 0.162          \\
                                & \mrs             & 0.017  & 0.054 & 0.078 & 0.099 & 0.118 \\
                                & \smodel (Ours)             & 0.018 & \textbf{0.044} & \textbf{0.067} & \textbf{0.089} & \textbf{0.110} \\



\bottomrule

\end{tabular}

\caption{Average Displacement Error Results for three Prisoner Escape (P-low, P-med, P-high) Datasets. Bolded values represent the best performing model.}
\label{tab:prisoner}
\vspace{-2mm}
\end{table}
We report our results on the three Prisoner Escape datasets in Table \ref{tab:prisoner}. We show that \smodel outperform or matches the prior baselines on all prediction horizons with the greatest advantages on the Prisoner-Medium and Prisoner-High datasets, \feedback{showcasing a 17\% and 12\% increase in predictive accuracy respectively}. We hypothesize that the flow-matching models are able to better incorporate the dense detection history information than the diffusion models because the flow field is deterministic and does not include adding an additional noise component. This may benefit the flow matching models to generate more confident and correct trajectories as compared to the diffusion models. 

We also show the ADE for the entire prediction horizon on the Prisoner-Medium dataset in Figure \ref{fig:cfm_result}. The VRNN and GRaMMI models are not shown as they do not predict full trajectories and also are not as competitive as the diffusion baseline. We find that our flow matching model reduces the ADE over all time horizons and has a tighter standard deviation than the diffusion model. Furthermore, we observe a performance dip in CADENCE between the first and second prediction timesteps, characterized by the sudden increase in ADE. This occurs because CADENCE employs an inpainting formulation that sets the first timestep to the detected location, if available. Consequently, this formulation introduces a risk of discontinuities in the diffusion tracks. Our results show that our flow matching model does not encounter the same issue and can outperform the diffusion-based model even without an explicit inpainting formulation.



    

\subsubsection{Sampling Speed Analysis}

We analyze the accuracy of our model compared to the diffusion model by reducing the total number of sampling steps $N$. In diffusion models, sampling steps refer to denoising steps, while in our method, they refer to Euler sampling steps. Both formulations require a neural network function call at each sampling step, making the number of sampling steps the primary bottleneck in reducing overall sampling time as the underlying neural network architecture is the same. 

\begin{table*}[tbp] \centering
\vspace*{1.5mm}
\begin{tabular}{@{}l|rrr|rrr|rrr@{}}
\toprule
& \multicolumn{3}{c}{Lon MAE}& \multicolumn{3}{c}{Lat MAE}& \multicolumn{3}{c}{Alt MAE} \\
& $0$ & $15$ & $30$ & $0$ & $15$ & $30$ & $0$ & $15$ & $30$ \\
\midrule
FlightBERT & 0.036 & 0.127 & 0.164 & 0.024 & 0.087 & 0.102 & 390.1 & 1060.8 & 1014.1 \\
T-CFM (Ours) & \textbf{0.010} & \textbf{0.098} & \textbf{0.130} & \textbf{0.006} & \textbf{0.067} & \textbf{0.075} & \textbf{145.3} & \textbf{853.3} & \textbf{782.6} \\
\midrule
& \multicolumn{3}{c}{Lon RMSE}& \multicolumn{3}{c}{Lat RMSE}& \multicolumn{3}{c}{Alt RMSE} \\
& $0$ & $15$ & $30$ & $0$ & $15$ & $30$ & $0$ & $15$ & $30$ \\
\midrule
FlightBERT & 0.057 & 0.188 & 0.267 & 0.035 & 0.122 & 0.161 & 509.9 & 1430.6 & 1375.1 \\
T-CFM (Ours) & \textbf{0.014} & \textbf{0.160} & \textbf{0.234} & \textbf{0.009} & \textbf{0.107} & \textbf{0.141} & \textbf{205.5} & \textbf{1242.6} & \textbf{1135.4} \\
\bottomrule
\end{tabular}
\caption{Aircraft Trajectory Forecasting: T-CFM achieves lower MAE and RMSE on Latitude, Longitude, and Altitude.}
\label{tab:aircraft_res}
\vspace{-4mm}
\end{table*}









We find that our flow-based model can generate high quality samples with just a single sampling step (Figure \ref{fig:adv_sample_speed}). This is due to the difference between flow matching and diffusion objectives. Flow matching enforces a straight probability flow between the starting distribution $q(\tau_0)$ and the ending distribution $q(\tau_1)$. Consequently, while diffusion models may need to adjust the sample direction during denoising, the flow matching framework learns a good initial estimate of how to move samples from the noisy distribution, enabling sample generation without multiple steps.
\subsection{Aircraft Trajectory Forecasting}

The trajectory forecasting task tests our model's generative capabilities on real-world data rather than simulated data. We compare our method against FlightBERT \cite{guo23flightbert}, a modern transformer-based framework built specifically for aircraft trajectory forecasting. We modify FlightBERT's attention mechanism, as the original framework assumed access to aircraft velocities. We also train with a negative log-likelihood loss to better model the variance in our dataset.

Table \ref{tab:aircraft_res} shows our model's performance as compared to FlightBERT. We find that our method outperforms FlightBERT on all metrics with an average improvement of $35.4\%$ over all metrics. We hypothesize two main reasons that our method outperforms FlightBERT. First, our method is not autoregressive generates the whole trajectory at once. This provides an advantage as errors may not accumulate over the prediction time horizon. FlightBERT was only tested for shorter horizon predictions. Meanwhile, we are interested in longer horizon predictions of up to 30 minutes compared to the 5 minute horizon for the dataset in FlightBERT. \feedback{Second, we hypothesize that the multimodal capabilities of our model is important for our flight trajectory dataset. Unlike commercial flights, the behavior of the Cessna aircraft does not travel in straight paths and consists of multiple heading changes throughout its path. We show that \smodel better models these diverse trajectories than prior work.}


\subsection{Long-Horizon Planning through Imitation Learning}

\begin{figure}[tbp]
\centering
    \includegraphics[width=0.85\columnwidth]{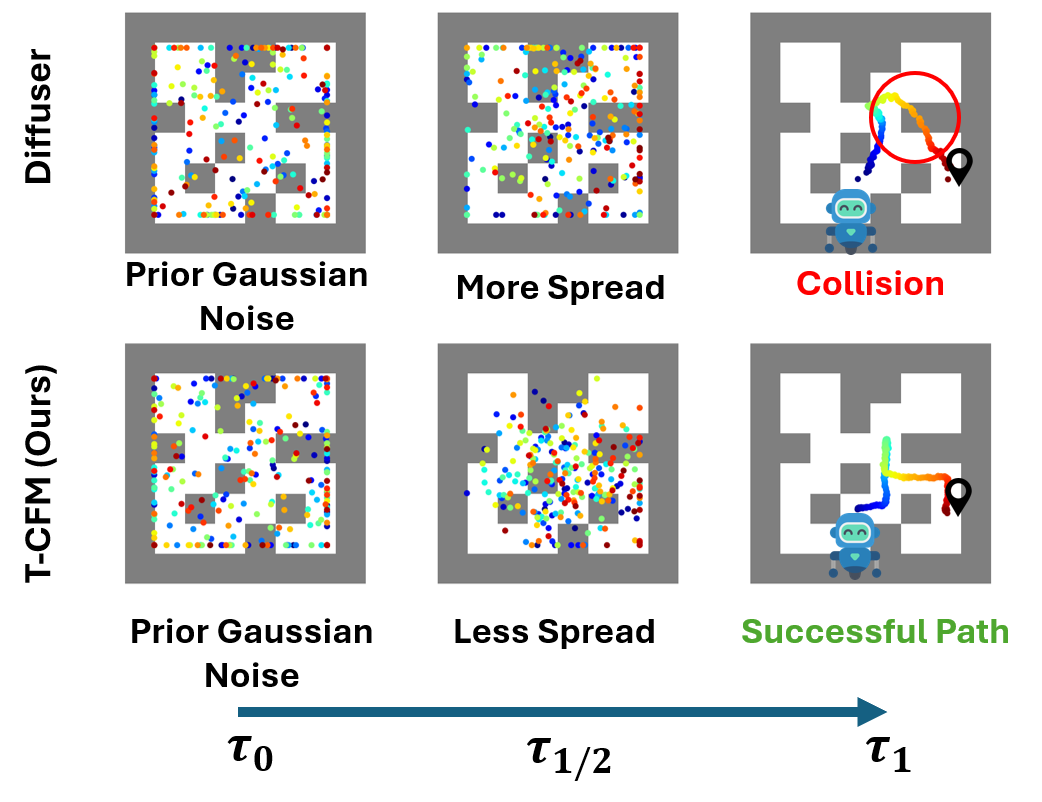}
    \caption{Visualization of Sampling Procedure between Diffuser (top) and \smodel (bottom) in Maze2D-Medium. In just two sampling steps, we show that \smodel can successfully plan a path between the start and end unlike Diffuser.}
    \label{fig:denoising_vis}
    \vspace{-5mm}
\end{figure}

We compare how well our models perform against Diffuser \cite{janner2022planning}, the current state-of-the-art method for learning how to plan solely from data on the Maze2D task (Table \ref{tab:maze2d-results}). We find that with just a single sampling step, our method significantly outperforms the diffusion models, \feedback{achieving a 142\% increase in score}. Similar to our analysis in Adversarial Tracking, the linear probability flows allows us to immediately infer high quality samples. \feedback{Additionally, we provide a visualization of the sampling procedure with just two sampling timesteps in Figure \ref{fig:denoising_vis}. Here we show that Diffuser produces a plan that intersects with the wall, as it is requires a large number of sampling steps to produce coherent plans. Meanwhile our method plans a collision free path. Additionally, the middle trajectory $\tau_{1/2}$ shows more coherence and has less spread in \smodel than Diffuser. This supports the hypothesis that our \model's flow field is more efficient at transforming noisy trajectories into realistic ones.} 

While \smodel outperforms Diffuser in the U-Maze and Medium mazes with more sampling steps, Diffuser outperforms our model in the Large maze. \smodel occasionally generates good plans but sometimes produce paths that collide with walls, lowering the overall score. Sampling a single trajectory from noise increases the chance of generating inaccessible plans compared to Diffuser. This may be because linear flows from flow matching struggle to correct certain noise initializations. In contrast, the diffusion model's sampling procedure is less dependent on initial noise, allowing it to reason better in the larger domain.

\begin{table}[ht]
\centering
\begin{tabular}{lllll}
\toprule
Environment & \multicolumn{2}{c}{N=1} & \multicolumn{2}{c}{N=256} \\
\cmidrule(lr){2-3} \cmidrule(lr){4-5}
& Diffuser & \multirow{-1}{*}{T-CFM} & Diffuser & \multirow{-1}{*}{T-CFM} \\
& & (Ours) & & (Ours) \\
\midrule
Maze2D U-Maze & 50.7\tiny{$\pm$}6.7 & \textbf{106.7}\tiny{$\pm$}2.7 & 112.5\tiny{$\pm$}11.2 & \textbf{122.1}\tiny{$\pm$}1.4 \\
Maze2D Medium & 21.7\tiny{$\pm$}13.5 & \textbf{112.2}\tiny{$\pm$}1.5 & 123.3\tiny{$\pm$}1.6 & \textbf{123.8}\tiny{$\pm$}3.5 \\
Maze2D Large & 30.3\tiny{$\pm$}6.9 & \textbf{111.0}\tiny{$\pm$}13.5 & \textbf{112.6}\tiny{$\pm$}16.3 & 104.3\tiny{$\pm$}3.4 \\
\midrule
Single-Task Average & 34.2 & \textbf{109.9} & 116.1 & \textbf{116.7} \\
\midrule
Multi2D U-Maze & 69.8\tiny{$\pm$}16.0 & \textbf{129.8}\tiny{$\pm$}3.0 & 127.3\tiny{$\pm$}3.3 & \textbf{129.5}\tiny{$\pm$}0.9 \\
Multi2D Medium & 58.4\tiny{$\pm$}5.3 & \textbf{116.5}\tiny{$\pm$}2.8 & 124.2\tiny{$\pm$}1.2 & \textbf{126.5}\tiny{$\pm$}4.1 \\
Multi2D Large & 35.8\tiny{$\pm$}4.2 & \textbf{121.7}\tiny{$\pm$}5.0 & \textbf{138.7}\tiny{$\pm$}5.9 & 127.3\tiny{$\pm$}8.6 \\
\midrule
Multi-Task Average & 54.7 & \textbf{122.7} & \textbf{130.1} & 127.8 \\
\bottomrule
\end{tabular}
\caption{The performance of \smodel and Diffuser on the long-horizon Maze2D compared when given a single sampling timestep $N=1$ and maximum sampling timesteps, $N=256$. Our model (T-CFM) is able to drastically reduce the number of sampling steps required to generate feasible plans whereas the Diffuser model fails at $N=1$.}
\vspace{-2mm}
\label{tab:maze2d-results}
\end{table}

\begin{figure}[htbp]
\centering
    \includegraphics[width=0.85\columnwidth]{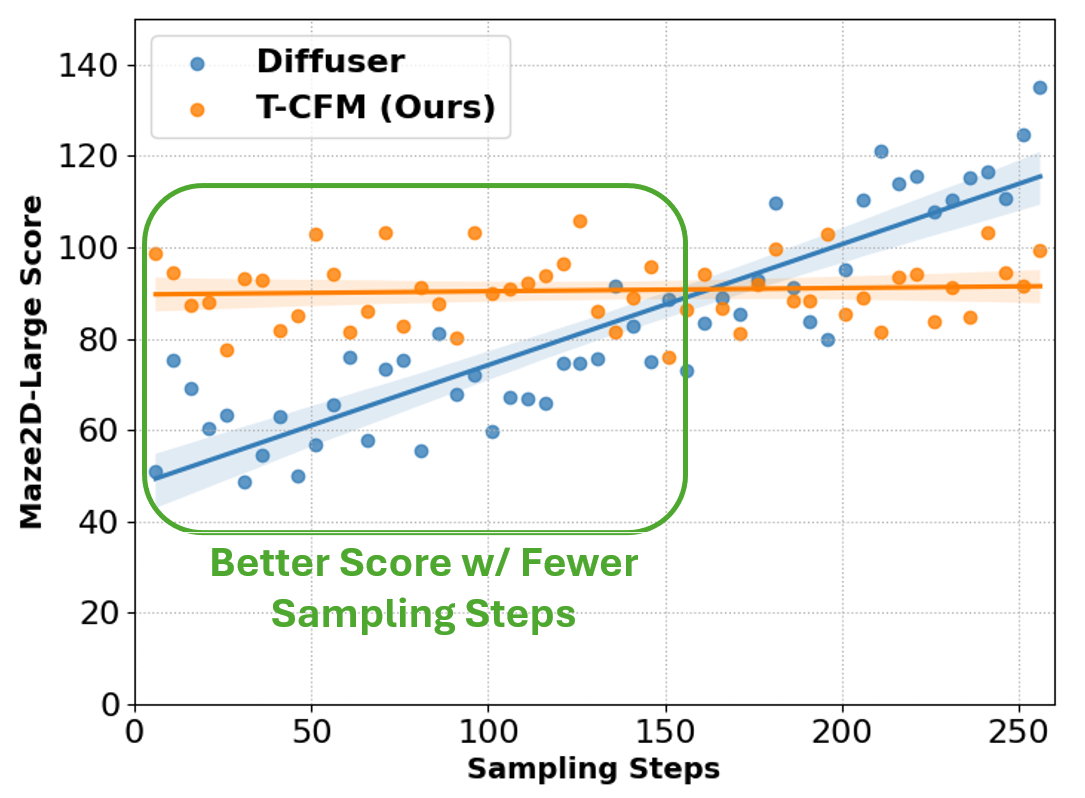}
    \caption{Compared to Diffuser, our T-CFM model does not drop in performance as we reduce the number of sampling steps on the Large Maze2D domain. }
    \label{fig:maze2d-result}
\vspace{-5mm}
\end{figure}

\section{Limitations and Future Work}
Our current approach does not explicitly consider multi-agent interactions. Future work includes extending to social navigation and autonomous driving scenarios, which require reasoning about other agents. We aim to increase our flow matching models' expressiveness, potentially incorporating stochastic bridge matching \cite{tong2023simulation} to combine deterministic ODE and stochastic SDE formulations. Further experiments with \smodel in dynamic situations will better demonstrate its capabilities for real-world robotic tasks.



\section{Conclusion}
\smodel is a novel approach for efficient trajectory forecasting and planning in robotics. By learning time-varying vector fields through flow matching, T-CFM achieves state-of-the-art performance on tasks like adversarial tracking, aircraft trajectory prediction, and long-horizon planning. T-CFM offers significant speed-ups compared to diffusion-based models without compromising accuracy, paving the way for more autonomous and responsive robots operating in complex, dynamic environments.

\bibliographystyle{plain}
\bibliography{references.bib}

\end{document}